\documentclass[10pt,journal,compsoc]{IEEEtran}

\ifCLASSOPTIONcompsoc
  \usepackage[nocompress]{cite}
\else
  \usepackage{cite}
\fi
\ifCLASSINFOpdf
\else
\fi

\usepackage{url}
\usepackage{times}
\usepackage{epsfig}
\usepackage{graphicx}
\usepackage{amsmath}
\usepackage{amssymb}
\usepackage{multirow}
\usepackage{colortbl}
\usepackage{color}
\usepackage{threeparttable}
\usepackage{booktabs}
\usepackage{adjustbox}
\usepackage{subfig}
\usepackage{caption}
\usepackage[misc]{ifsym} 

\newcommand{\RR}{\mathbb{R}}

\begin{document}
%
\title{Improving Video Instance Segmentation via Temporal Pyramid Routing}
%
%
%
%

\author{Xiangtai Li,
        Hao He,
        Yibo Yang,
        Henghui Ding,
        Kuiyuan Yang,
        Guangliang Cheng,\\
        Yunhai Tong \textrm{\Letter},
        Dacheng Tao,~\textit{Fellow}, IEEE
        
\IEEEcompsocitemizethanks{\IEEEcompsocthanksitem X. Li and Y. Tong are with the School of Electronics Engineering and Computer Science, Peking Univeristy, Beijing, China. This work is supported by the National Key Research and Development Program of China (No.2020YFB2103402). 
\IEEEcompsocthanksitem H.~He is with the Department of National Laboratory of Pattern Recognition, Institute of Automation, Beijing , China.
\IEEEcompsocthanksitem Y.~Yang and D.~Tao are with JD Explore Academy, Beijing, China.
\IEEEcompsocthanksitem G.~Cheng is with SenseTime Research, Beijing, China.
\IEEEcompsocthanksitem H.~Ding is with ETH Zurich, Switzerland.
\IEEEcompsocthanksitem K.~Yang is with Xiaomi, Beijing, China.
}
}

%
%

\markboth{IEEE TRANSACTIONS ON PATTERN ANALYSIS AND MACHINE INTELLIGENCE,~Vol.~X, No.~X, X}
{Shell \MakeLowercase{\textit{et al.}}: Bare Advanced Demo of IEEEtran.cls for IEEE Computer Society Journals}
%



\IEEEtitleabstractindextext{
\begin{abstract}
Video Instance Segmentation (VIS) is a new and inherently multi-task problem, which aims to detect, segment, and track each instance in a video sequence. Existing approaches are mainly based on single-frame features or single-scale features of multiple frames, where either temporal information or multi-scale information is ignored. To incorporate both temporal and scale information, we propose a Temporal Pyramid Routing (TPR) strategy to conditionally align and conduct pixel-level aggregation from a feature pyramid pair of two adjacent frames. Specifically, TPR contains two novel components, including Dynamic Aligned Cell Routing (DACR) and Cross Pyramid Routing (CPR), where DACR is designed for aligning and gating pyramid features across temporal dimension, while CPR transfers temporally aggregated features across scale dimension. Moreover, our approach is a light-weight and plug-and-play module and can be easily applied to existing instance segmentation methods. Extensive experiments on three datasets including YouTube-VIS (2019, 2021) and Cityscapes-VPS demonstrate the effectiveness and efficiency of the proposed approach on several state-of-the-art instance and panoptic segmentation methods. Codes will be publicly available at \url{https://github.com/lxtGH/TemporalPyramidRouting}.
\end{abstract}

\begin{IEEEkeywords}
Dynamic Network, Temporal Modeling, Video Instance Segmentation, Video Scene Understanding
\end{IEEEkeywords}}

\maketitle

\IEEEdisplaynontitleabstractindextext

%
\IEEEpeerreviewmaketitle

\ifCLASSOPTIONcompsoc
\IEEEraisesectionheading{\section{Introduction}\label{sec:intro}}
\else
\section{Introduction}
\label{sec:intro}
\label{sec:introduction}
\fi



 
 \IEEEPARstart{M}{odeling} dynamic video inputs for scene understanding is a fundamental vision research direction~\cite{vis_dataset,kim2020vps}. Video Instance Segmentation (VIS)~\cite{vis_dataset}, has been proposed and drawn lots of attention. VIS aims to simultaneously classify, segment and track object instances in a given video sequence, which is a very challenging task that involves the instance correlations along the temporal dimension.

 \begin{figure}[!t]
	\centering
	\includegraphics[width=0.85\linewidth]{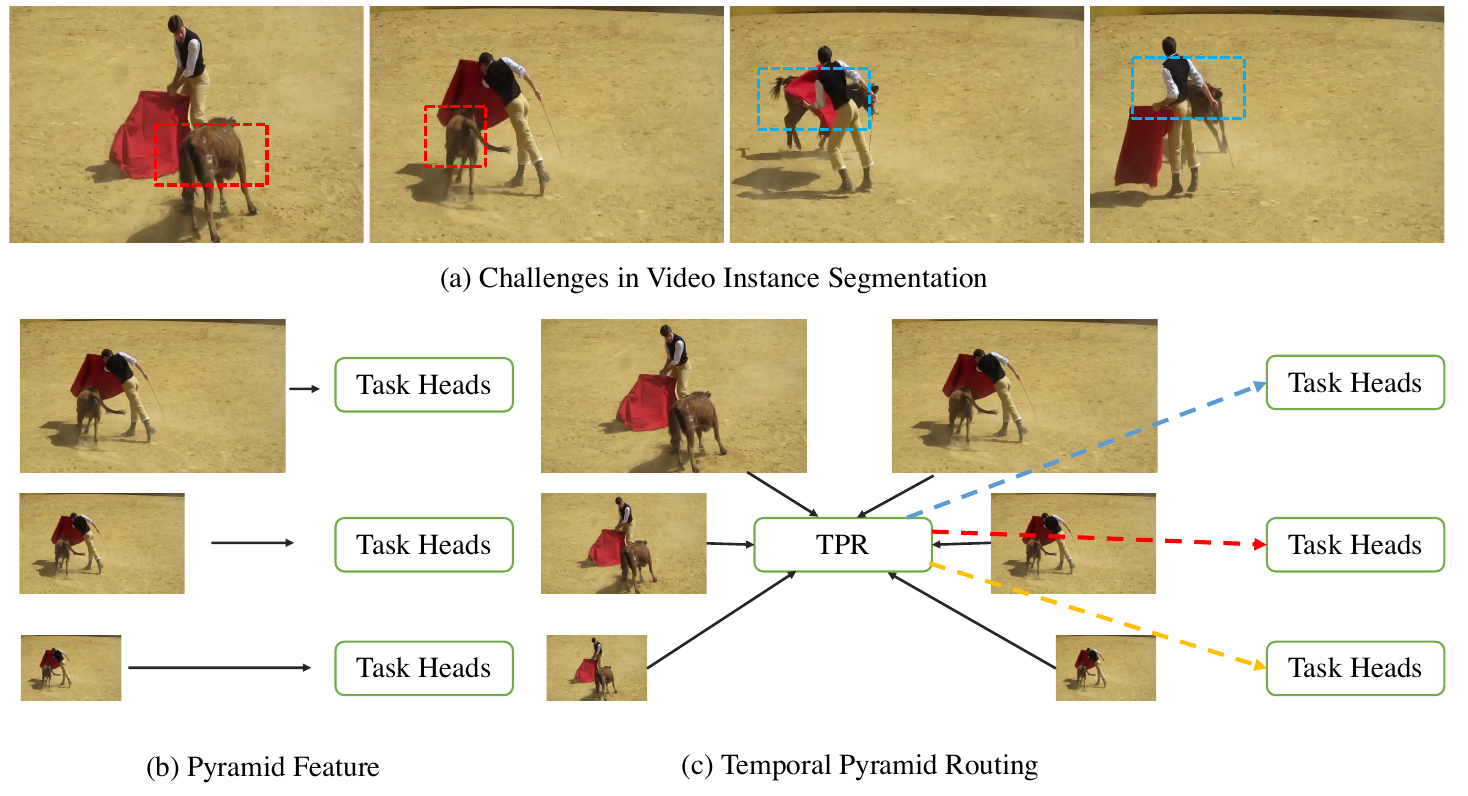}
	\caption{(a) The illustration of problems in video instance segmentation: scale variation (red boxes) and overlapping (blue boxes). (b) Pyramid feature learning in image understand tasks. (c) Our proposed Temporal Pyramid Routing. Each line in a different color represents per pixel-wise routing across different scales along the temporal dimension.
	}
	\label{fig:teaser}
\end{figure}

Previously, many state-of-the-art methods~\cite{mask_pro_vis,VIS_TR} model the temporal correlation by feature aggregation. For example, Mask-Prop~\cite{mask_pro_vis} proposes to warp mask features via a strong object detector~\cite{htc} while VisTR~\cite{VIS_TR} treats the entire video clips as sequence inputs and aggregates features via the transformer detector~\cite{detr}. Moreover, bottom-up method~\cite{Athar_Mahadevan20ECCV} adopts separate object instances by clustering learned pixel embeddings~\cite{neven2019instanceSeg} then performs tracking and association via specific post-processing. However, these methods pay little attention to the scale variation problems when modeling temporal feature aggregation, resulting in ambiguous predictions for large variation objects. As shown in Fig.~\ref{fig:teaser}(a), the cow in one video clip has a large-scale variation between frames. To segment and track objects consistently across consecutive frames in a video, it is essential to incorporate past context and feature pyramids through a temporal model. The scale variation problems have been well studied in the image domain via pyramid representation, as shown in Fig.~\ref{fig:teaser}(b). Feature Pyramid Network (FPN)~\cite{fpn,focal_loss,maskrcnn} and its variants~\cite{PANet,nasfpn,tan2020efficientdet} are the most classic architectures to establish a pyramid network for object representation. It assigns instances to different pyramid levels according to the object sizes. However, it has \textit{not} been well explored in the video on how to align other feature pyramids among the different feature levels along the temporal dimension. Also, it has not been well studied on how to model the spatial-temporal feature pyramids for the down-stream tasks. 

Several works utilize optical flow~\cite{DFF,zhu2017flowagg} or non-local operators~\cite{chen2020mega} to aggregate temporal features for video object detection. However, both limited scale variation and simple scenario on ImageNet VID dataset~\cite{russakovsky2015imagenet} make current methods mainly focus on single scale feature aggregation using Faster R-CNN~\cite{ren2015faster} as the baseline. In this work, we aim to design an approach that enhances the network's ability to handle the video domain's scale variation problem.

Directly fusing the previous feature pyramid into the current corresponding feature pyramid is a trivial solution because there may be significant motion variation and background noise in the video. Thus, filtering irrelevant features out and dynamically assigning relevant parts in the previous frame is essential for temporal consistent feature representation. However, existing warping-based approaches~\cite{zhu2017flowagg,zhu2018towards,chen2020mega} may lead to high computation costs. Our method is inspired by the success of dynamic routing in image tasks~\cite{song2020_dynamic_head,li2020learningDynamicRoutingSeg}. It performs the pixel-wised fusion across the feature pyramid, controlled by adaptive learned gates. Compared with static network, it has the advantage of modeling various inputs since the network architecture can be changed according to the inputs.  We extend such dynamic network into the video scene understanding and propose a conceptually novel method for temporal feature learning, called Temporal Pyramid Routing (TPR). It dynamically aligns and fuses features, where the pixel-wise routing is performed between adjacent feature pyramids. As shown in Fig.~\ref{fig:teaser}(c), our TPR takes two adjacent temporal feature pyramids as input and outputs refined feature pyramids for the down-stream task.

\begin{figure}[!t]
	\centering
	\includegraphics[width=0.90\linewidth]{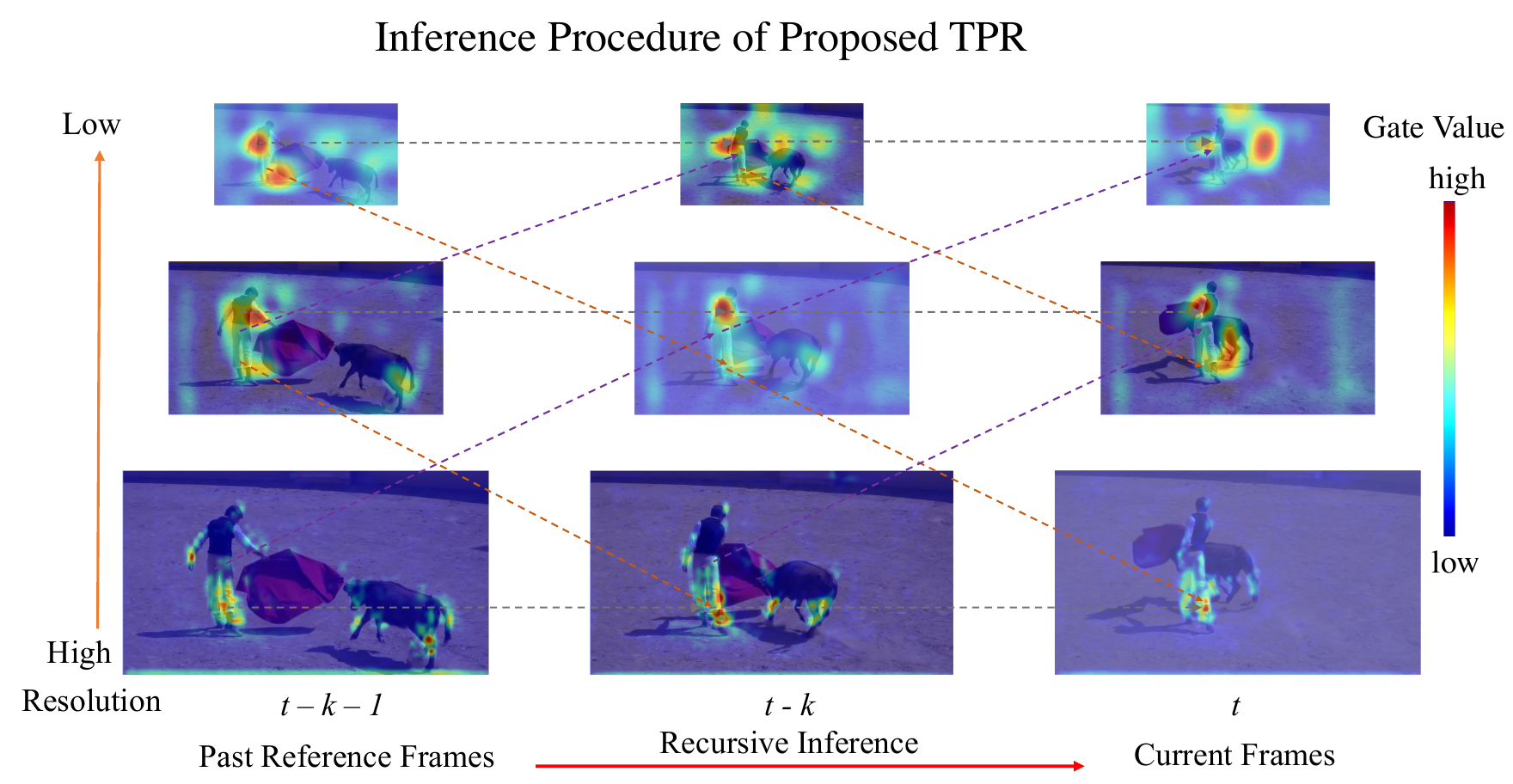}
	\caption{ An illustration of the inference procedure of our proposed TPR. \textbf{Only parts of the feature} (highlighted by gates) are propagated into the next frame, \textit{sparsely} and \textit{recursively}.}
	\label{fig:teaser_inference}
\end{figure}

In particular, we design a novel pyramid routing space that mainly contains two steps: Dynamic Aligned Cell Routing (DACR) and Cross Pyramid Routing (CPR). The former uses specifically designed double gates to filter out background noise and absorb relevant semantics. The latter introduces the aligned features from the previous frame into the remaining scales of the current frame dynamically and efficiently. Fig.~\ref{fig:teaser_inference} gives a visual example of the TPR inference procedure. For each pyramid feature, only highlight parts from the previous frame are propagated into the current frame via pixel-level routing, which avoids the complex background noise and redundant information. Finally, we verify our proposed TPR on various baselines~\cite{chen2020blendmask,Cao_SipMask_ECCV_2020,yolact-iccv2019} on the YouTube-VIS dataset~\cite{vis_dataset}. The experimental results show that our proposed TPR improves baselines by a significant margin (about 2\%-3\% mAP) with little extra GFlops. Overall, the proposed TPR is fundamentally different from the existing feature aggregation methods in video. TPR has utilized pixel-wise gated routing while maintaining inference efficiency, while previous works~\cite{chen2020mega,zhu2017flowagg} introduce much more computation and redundant information between the frames. With Swin Transformer~\cite{liu2021swin} as backbone, our method can achieve 45.9 mAP (online inference) which outperforms MaskProp~\cite{mask_pro_vis} with~\textit{faster} inference speed (about 10 time faster than MaskProp). Our main contributions are summarized as follows:

1) We propose a novel framework named Temporal Pyramid Routing for temporal dynamic multiscale representation learning. In particular, we design two main components, including Dynamic Aligned Cell Routing and Cross Pyramid Routing to aggregate features from the previous frame while keeping efficiency.
2) Our TPR is a plug-in module where we append it into several image segmentation methods to perform temporal dynamic routing for two different video segmentation tasks, including video instance segmentation and video panoptic segmentation.
3) Detailed experiments and analyses on the YouTube-VIS dataset indicate our proposed TPR's efficacy in improving the accuracy and keeping light-weight. Our method improves baselines by a significant margin. Our method achieves the state-of-the-art results \textit{under the online inference setting}. We further verify the generality of TPR on Video Panoptic Segmentation (VPS). 

\section{Related Work}

\noindent
\textbf{Instance Segmentation} Instance Segmentation aims to detect and segment each instance~\cite{dai2016instance,simultaneous_det_seg}. The two-stage pipeline Mask-RCNN and its variants~\cite{maskrcnn,msrcnn,htc} first generate object proposals using Region Proposal Network (RPN)~\cite{ren2015faster} and then predict boxes and masks on each RoI feature. Recently, several single-stage methods~\cite{tian2020conditional,chen2020blendmask,yolact-iccv2019} achieve significant progress and comparable results with two-stage pipelines. Meanwhile, there are several bottom-up approaches~\cite{de2017semanticInstanceLoss}. Our method is built on those top-down works and explores the correlation of the temporal feature pyramid.

\noindent
\textbf{Video Instance Segmentation} The VIS requires classifying, segmenting each instance for each frame, and assigning the same instance with the same id. MaskTrack-RCNN~\cite{vis_dataset} is the first attempt to address this task, where they also propose YouTube-VIS dataset for benchmarking video instance segmentation algorithms. Maskprop~\cite{mask_pro_vis} is built on state-of-the-art detection method HTC~\cite{htc} and crops the extracted features via predicted masks, then propagates them temporally to improve the segmentation and tracking. Based on~\cite{vis_dataset}, the work~\cite{lin2020video_vae} proposes a modified variational autoencoder, while CompFeat~\cite{fu2021compfeat} aggregates both frame-level and object-level information. Recently, VisTR~\cite{VIS_TR} solves VIS problem using detection transformer~\cite{detr}. VIS is also closely related to Video Semantic Segmentation~\cite{DFF} and Video Panoptic Segmentation~\cite{kim2020vps}. Multi-Object Tracking and Segmentation (MOTS) task~\cite{voigtlaender2019mots} is proposed to evaluate MOT along with instance segmentation. Recently, several works~\cite{video_knet,yuan2021polyphonicformer,zhou2022transvod} use object query to unify video scene understanding tasks. Due to its limited scale distribution and much fewer object categories, we do not compare it in this paper. TPN~\cite{yang2020temporal} also proposes to use multiscale temporal pyramid networks for action recognition. However, this method cannot generalize to video dense prediction tasks due to the misaligned temporal representation. 
 
\noindent
\textbf{Dynamic Network Design} Several works design the specific dynamic components~\cite{yang2019condconv,chen2020dynamic_conv,DynamicCapacityNetwork} and architectures~\cite{huang2017multi} for image classification task where the parameters and architectures are conditioned on inputs. Recently, inspired by Neural Architecture Search (NAS)~\cite{liu2018darts,cai2018proxylessnas,pham2018efficient,zoph2016nas}, several works propose dynamic network design on the dense pixel prediction tasks such as semantic segmentation~\cite{li2020learningDynamicRoutingSeg} and object detection~\cite{song2020_dynamic_head}. These methods have data-dependent routes which are adapted from the scale distribution of each input. However, the dynamic network for video dense prediction tasks including VIS and VPS are \textit{not well explored}. Our methods are inspired by those works, but differ from them. In particular, we design several new routing cells to align and fuse temporal information.
Moreover, a new routing space is introduced to perform cross fusion between feature pyramids. To the best of our knowledge, we are the first to design a novel temporal routing strategy for the video segmentation tasks. Moreover, our methods are \textit{fundamentally different} from previous attention based approaches~\cite{woo2018cbam} where the computation cost is fixed during inference. Our dynamic architecture propagates parts of features according to the input, which has been shown effective and robust. 

\section{Method}


\subsection{Overview of Dynamic Routing in Image Domain}
\label{sec:overview}

\begin{figure*}[!h]
	\centering
	\includegraphics[width=0.90\linewidth]{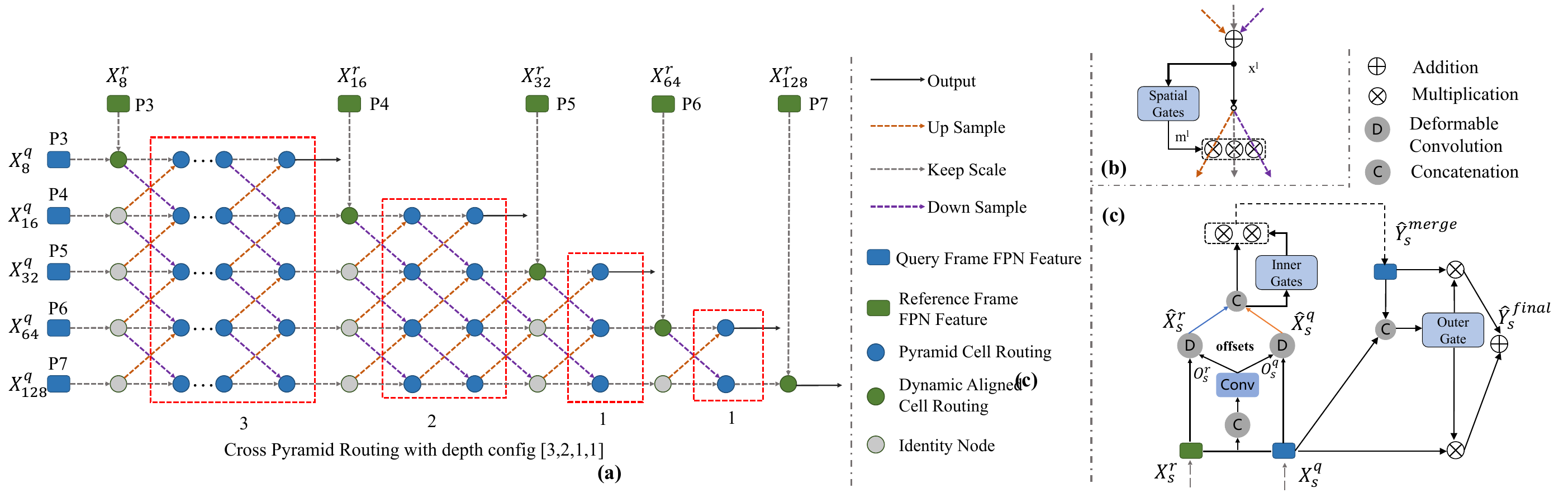}
	\caption{ An illustration of our proposed Temporal Pyramid Routing system, which takes two feature pyramids as inputs (blue rectangles and green rectangles), and outputs the refined feature pyramids for down-stream tasks. Following VIS paper~\cite{vis_dataset}, Query frame represents the current frame while Reference frame is randomly sampled from the same video clip around the current frame. (a) Overview of our TPR. It mainly contains two components: Dynamic Aligned Cell Routing (Green Circles) and Cross Pyramid Routing (Blue Circles in Red dotted boxes). (b) Detailed cell operation in Cross Pyramid Routing. (c) Detailed cell operation in Dynamic Aligned Cell Routing.
	}
	\label{fig:methods}
\end{figure*}

\noindent
\textbf{Basic Notation and Concepts.}
Compared with static networks, dynamic networks~\cite{chen2020dynamic_conv,li2020learningDynamicRoutingSeg,song2020_dynamic_head} have larger network capacity and better results under certain budget constraints. In a dynamic network, a \emph{routing space} is defined as \emph{all potential paths}, and the routing process is to aggregate potential multiscale features and choose routing paths conditioned on each input image using \emph{cells} and \emph{gates} in each path. \emph{Cells} are the basic computation blocks, while the \emph{gates} control how much computation is performed conditionally for each input. Formally, given the input feature $X_{s}^{l} \in \RR^{ C \times H \times W}$ with $H\times W$ pixel-level locations and $C$ channels, a cell processes the feature with an operation set $\mathcal{O}$ contains widely-used network operators (e.g., convolutions, identity mapping, pooling),
\begin{equation}\label{equ:cell_inside}
{H}_s^l=\sum_{O^i\in \mathcal{O}}O^i({X}_s^l),
\end{equation}
where  $l$ means the current depth of routing space, $s \in \{ 2^{i}, i=3,4,...,k\}$ denotes the feature scale, and $k$ is the largest downsample ratio, for example, in FCOS~\cite{tian2021fcos}, $k$=7, namely P3, P4, ... ,P7. 
The output ${H_{s}^{l}}$ is transferred to different scales according to the gating factor ${m_{s}^{l}}$. Take pixel-wise routing gates as an example, each pixel-level location has a gate factor generated by 
\begin{equation}\label{equ:gates}
{m}_s^l=G_{s}^l({X}_s^l;\theta_{s}^{l}),
\end{equation}
where $\theta_{s}^{l}$ is the network parameters learned for defining the gating function $G_{s}^l(\cdot)$.
The output ${m_{s}^{l}}$ is normalized via a variant of the soft differentiable tanh function~\cite{song2020_dynamic_head}. Finally, $X_{s}^{l}$ is transformed into $Y_{s}^{l}$ after going through a path formed by cells and gates, i.e.,
\begin{equation}\label{equ:cell_output}
{Y_{s}^{l}} = {H_{s}^{l}} \cdot {m}_{s}^{l}.
\end{equation}

In summary, Dynamic Routing controls the operators applied on an image via gates conditioned on the image content, which is different from NAS-based approaches~\cite{liu2018darts,zoph2016nas}, where the network structure is only varied during training phase but fixed for inference.

\noindent
\textbf{Motivation for Video Instance Segmentation.} 
As shown in Fig.~\ref{fig:teaser}(a), a video clip contains a sequence of images stacked along the temporal dimension. Compared with a static image, a video contains more but redundant information, and also more variations caused by motion. Conditioned on a video input, dynamic routing is a natural mechanism to select useful instance information for VIS.

\subsection{Temporal Pyramid Routing}
\label{sec:tpr}

\noindent
\textbf{Temporal Pyramid Dynamic Routing Space.}
The basic task for dynamic network is routing space design, which determines the potential capacity of the network. In this work, we use feature pyramid to represent multiscale information of each frame, and feature pyramids from a video to capture temporal information. The routing space is designed to control the information flow along both the temporal and scale dimensions. Feature pyramids go through the routing space, and are conditionally refined by selected paths. The whole routing process is defined as Temporal Pyramid Routing (TPR). Finally, task-specific heads are attached to the refined feature pyramid for down-stream tasks, such as detection, segmentation and tracking. To be noted, it is a plug-and-play module that can be applied to any method using feature pyramid as input, which is verified in Sec.~\ref{sec:youtubevis}. 

As shown in Fig.~\ref{fig:methods}(a), we take two feature pyramid inputs in one video sequence as an example. One feature pyramid $X_{s}^{q}$ is from the query frame $q$ (i.e., current frame, denoted as blue rectangles) and the other feature pyramid $X_{s}^{r}$ is from the reference frame $r$ (i.e., support frame, denoted as green rectangles). The two input feature pyramids go through TPR, which routes temporal information via Dynamic Aligned Cell Routing (DACR) and scale information via Cross Pyramid Routing (CPR). DACR absorbs the most relevant information from the reference frame in a pixel-wised manner scale by scale and CPR routes the absorbed information to other scales. We will specify the designing details of these two components.

\noindent
\textbf{Dynamic Aligned Cell Routing.} As mentioned above, it is crucial to design a temporal pixel-wise routing cell to avoid noises and sample the most important parts. We apply a sample and filter strategy to align temporal feature frames. First, Dynamic Aligned Cell Routing utilizes the dynamic sampling operator~\cite{dai2017deformable} to sample corresponding pixel locations, and then irrelevant parts are filtered out via our double gates design. Two features $X_{s}^{q}(i)$ and $X_{s}^{r}(i)$ are the inputs of our Dynamic Aligned Cell, where $i$ is the index of each spatial position. The first step is to find the correspondence between the two features. We concatenate two features and generate two individual offset fields $O_{s}^{r}, O_{s}^{q}$ via one $3 \times 3 $ convolution layer. Then the corresponding features are sampled according to the predicted offset fields through deformable convolution (DCN)~\cite{dai2017deformable}. Take the query feature $X_{s}^{q}(i)$ with predicted offset $O_{s}^{q}$ as an example, 
\begin{equation}\label{equ:deformable_sample}
{\hat{X}_{s}^{q}(i)}= \sum_{p_{n} \in R} {X_{s}^{q}}(i+p_n+ O_{s}^{q}(p_{n})) \cdot W(p_{n}),
\end{equation}
where $p_{n}$ enumerates the locations in neighbor $R$ and $W$ is kernel weight of DCN. Then we perform pixel-wise dynamic routing using refined ${\hat{X}_{s}^{q}}$, ${\hat{X}_{s}^{r}}$ and original $X_{s}^{q}$ via our double gates. 

There are two types of gate in DACR, i.e., inner gate and outer gate. The inner gate aims to control the sampled reference features and find fine-grained supports from reference features, while the outer gate emphasizes informative locations in query frame and only absorbs features from reference frames accordingly. In particular, for inner gates, we choose stride convolution and max-pooling for gate map generation to highlight the most salient areas. Following Equ.~\ref{equ:cell_output}, the output of inner cell routing is formally defined as,

\begin{equation}\label{equ:dynaic_inner_cell_output}
{\hat{Y}_{s}}= \{ {Y_{s}^i} | {Y_{s}^i} = {H_{s}^i} \cdot Gate(Cat({\hat{X}_{s}^{q}}, {\hat{X}_{s}^{r}})), i \in{q, r} \},
\end{equation}
\noindent
where $Cat$ means concatenation operation, $Gate$ is the inner function and the hidden states ${ {H_{s}^i} }$ are generated by cell via Equ.~\ref{equ:cell_inside} where the inputs are the aligned outputs from Equ.~\ref{equ:deformable_sample}. During inference, only pooled areas that are selected by the inner gate can be propagated into ${\hat{Y}_{s}^{q}}$, which can be efficiently implemented via masked convolution~\cite{2014SpatiallyMaskConv}. Finally, we fuse two features ${Y_{s}^q}$ and ${Y_{s}^r}$ via $1 \times 1$ convolution, and denote the fusion result as $Y_{s}^{merge}$ for short. 

Although the inner gates can be effective to avoid background noises, the contents are flexibly dependent on sampled offsets $O_{s}^{r}, O_{s}^{q}$, which affect the stability of the query frame feature learning. Thus, we propose to combine query frame feature $X_{s}^{q}$ and merged feature $Y_{s}^{merge}$ via an outer gate map $G_{s}$. The gate map $G_{s}$ is generated by convolutional fusion of  $X_{s}^{q}$ and $Y_{s}^{merge}$ followed by a sigmoid function. The convolutional fusion is simply implemented by two $1 \times 1$ convolution. Then, the final feature map is obtained as the weighted sum of $X_{s}^{q}$ and $Y_{s}^{merge}$ according $G_{s}$,
\begin{equation}\label{equ:dynaic_out_cell_output}
{\hat{Y}_{s}^{final}}= G_{s} \cdot X_{s}^{q} + (1 - G_{s}) \cdot Y_{s}^{merge}
\end{equation}

\begin{figure}[!t]
	\centering
	\includegraphics[width=0.90\linewidth]{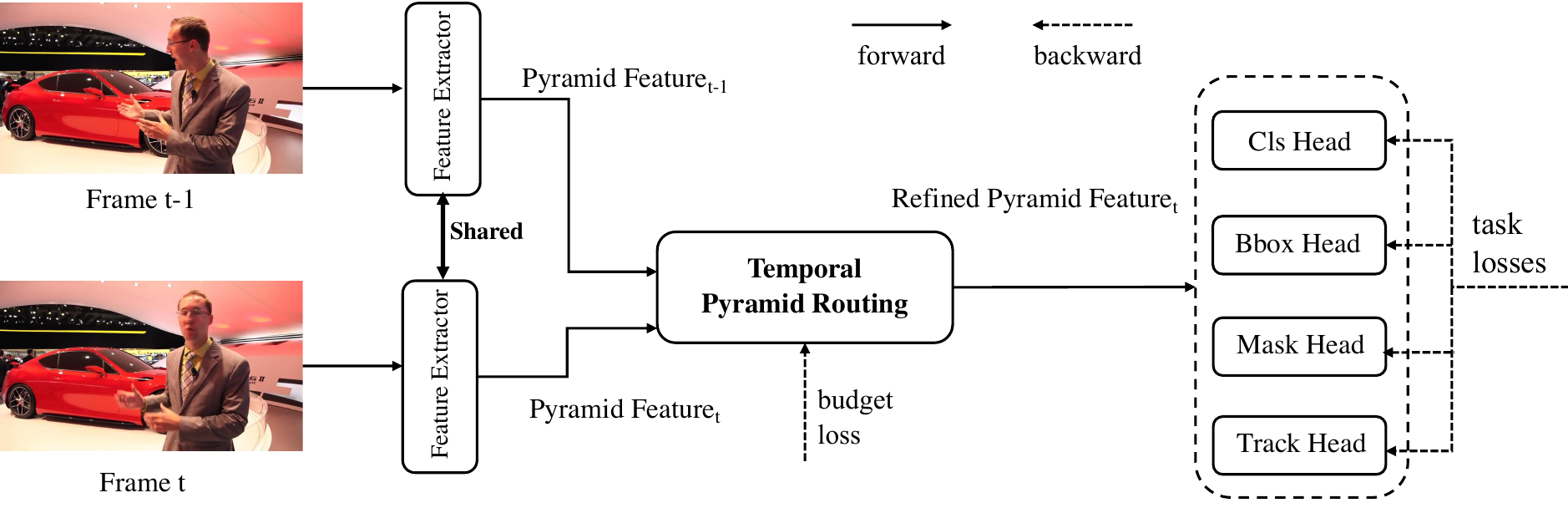}
	\caption{ An illustration of network architecture with our proposed TPR. The feature extractor contains backbone and FPN. During inference, Frame $t-1$ is used as reference frame. Best view it on screen. 
	}
	\vspace{-3mm}
	\label{fig:network}
\end{figure}

\noindent
\textbf{Cross Pyramid Routing.}
After harvesting reference features from each scale $s$, the next important problem is how to distribute ${\hat{Y}_{s}^{final}}$ to other scales of query feature pyramid. In FPN-like networks~\cite{fpn,PANet}, semantically strong features in low-resolution and semantically low features in high-resolution are combined in the network, which are also used in most state-of-the-art image instance segmentation approaches~\cite{chen2020blendmask,yolact-iccv2019}. CPR follows the  cross-scale, pixel-wise routing strategy~\cite{song2020_dynamic_head} with two exceptions. One is the input feature that is modified by DACR. The other is the lower-triangular shaped routing space, where lower-level features go through deeper paths to upper-level features, to bridge the semantic gap. The default routing depth $l$ is $[3,2,1,1]$ is used in all experiments, more choices about $l$ can be found in the supplementary. The detailed procedure of each routing cell is shown in Fig.~\ref{fig:methods}(b). Both DACR and CPR work at the same time for both training and inference.

\begin{figure}[!t]
	\centering
	\includegraphics[width=0.95\linewidth]{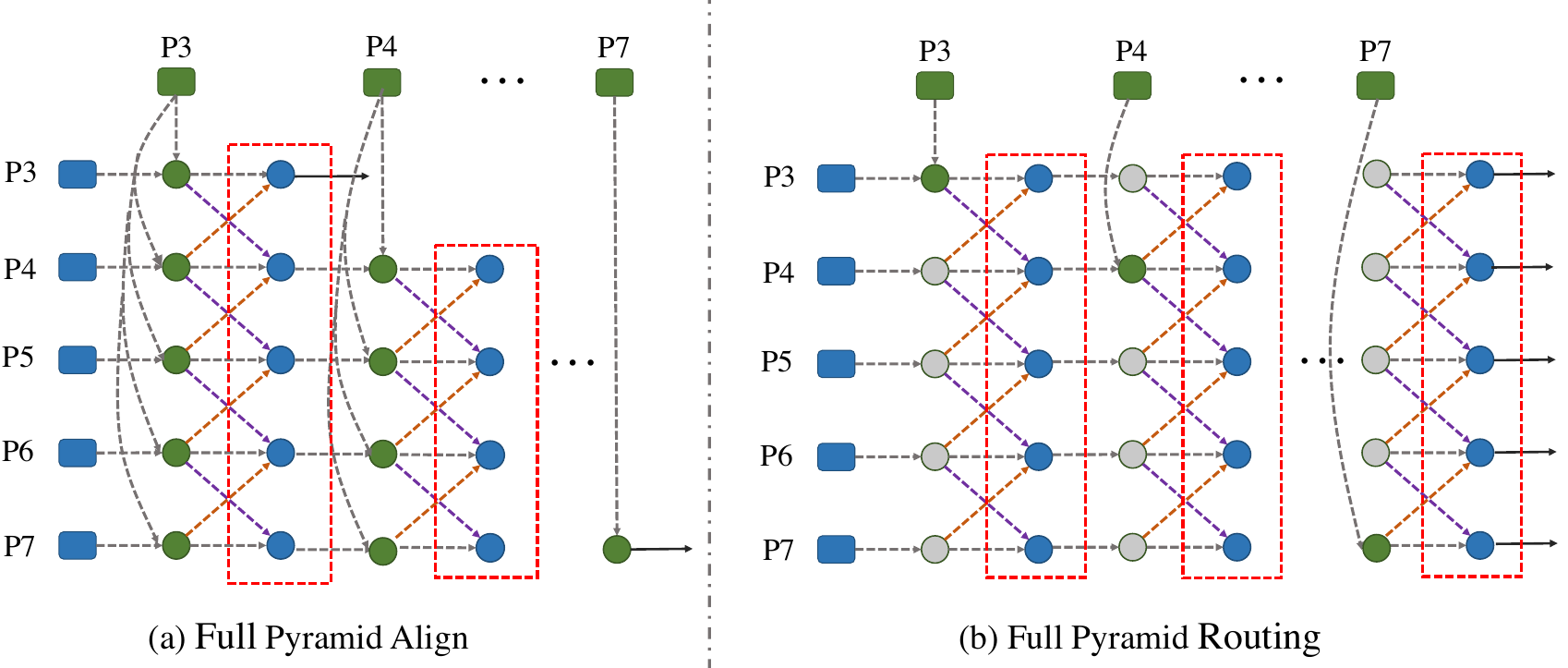}
	\caption{ An illustration of more routing space designs. (a). Full pyramid align. (b). Full pyramid routing.
	}
	\vspace{-3mm}
	\label{fig:align_compare}
\end{figure}

In addition to the proposed CPR design, there are another two straightforward design choices. One is Full Pyramid Align routing as illustrated in Fig.~\ref{fig:align_compare}(a), where each node from the reference feature pyramid is aligned to all nodes in the query feature pyramid via DACR. The other is Fully Pyramid Routing as proposed in~\cite{li2020learningDynamicRoutingSeg}, where all nodes in query feature pyramid are with the same routing depth. The former suffers from the severe misalignments between features across different scales. The latter treats all scales equally and leads to more redundant computation for upsampling high-level features.
Moreover, the direction of CPR is \textbf{not} top-down. There are three reasons: Firstly, adopting bottom-up routing for CPR has better results due to the complementary property of FPN (top-down). Secondly, it is also beneficial for aligned high resolution representation propagation, since misalignment in low resolution is severer~\cite{2016Clockwork}. Thirdly, bottom-up routing also achieves faster inference speed since high resolution features go out in the earlier stages. More detailed results can be found in Sec.~\ref{sec:youtubevis}.

\noindent
\textbf{Discussion With Warping Based Approaches.} Compared with previous works using optical flow for warping features~\cite{DFF,zhu2017flowagg}, our dynamic network does not require the extra optical flow network training and adaption. Compared with DCN sampling~\cite{bertasius2018object}, our proposed DACR can avoid background noises and achieve faster inference speed since only sparse locations are involved during the inference.  

\subsection{Network Architectures} 
\label{sec:nework}  
\noindent
\textbf{Extending Image Instance Segmentation into Video} VIS requires the association of instances in a video. Following previous works~\cite{vis_dataset,Cao_SipMask_ECCV_2020}, we also add a tracking head for fair comparison, as shown in Fig.~\ref{fig:network}. The tracking head outputs feature embedding of each instance candidate. Suppose there are $K$ instances identified from previous frames, the candidate box $i$ in the current frame will be assigned to a label $k$ according to the assignment probability,

\begin{equation} \label{eq:assignment_prob}
p_{i}(k)=
\begin{cases}
\frac{e^{f_{i}^{T}f_k}}{1+\sum_{j=1}^{K}e^{ f_{i}^{T}f_j }}& k\in{[1,K]},\\
\frac{1}{1+\sum_{j=1}^{K}e^{ f_{i}^{T}f_j }}& k=0,
\end{cases}
\end{equation}
where $1\leq k\leq K$ indicates the candidate is associated to one of the $K$ instances and $k=0$ means the candidate is treated as a new track, $f_i$ and $f_j$ ($j\in{[1,K]}$) denotes the feature embedding of the candidate and the $K$ pre-identified instances, respectively. Then we perform the cross-entropy loss by treating it as a multi-class classification problem. As shown in Fig.~\ref{fig:network}, we insert our proposed TPR between feature extractor and these task-specific heads (detection, segmentation, tracking) for VIS. 

\noindent
\textbf{Loss Functions} Since the dynamic network has a large capacity while we only have limited computational resources, we constrain the computational cost following previous works~\cite{li2020learningDynamicRoutingSeg,song2020_dynamic_head}. We consider all the locations involved in the receptive field of locations with positive gating factors, which are obtained by a max-pooling layer. We denote such computation budget from that gate as $B^{l}$ for the layer $l$. The budget loss is normalized by the overall computational complexity $C^{l}$.

\begin{equation}\label{equ:budget}
    \mathcal{L}_\mathrm{budget} = \frac{\sum_{l}B^{l}}{\sum_{l}C^{l}}
\end{equation}

\noindent
Finally, the total loss $\mathcal{L}$ is the combination of tasks-specific heads loss and budget loss. The positive hyper-parameters $\lambda_1$ and $\lambda_2$ are set to achieve a trade-off between efficiency and effectiveness, where we set $\lambda_1=1$ and $\lambda_2=1.5$ following~\cite{song2020_dynamic_head} by default. The $\mathcal{L}_\mathrm{tasks}$ contains detection, segmentation and tracking loss. In our experiments, we find that increasing $\lambda_1=1$ hurt the performance since most features from the previous frame are not aligned into current frame during inference. 

\begin{equation}\label{equ:final_loss}
\mathcal{L}=\lambda_1\mathcal{L}_\mathrm{tasks}+\lambda_2\mathcal{L}_\mathrm{budget}.
\end{equation}

\noindent
\textbf{Training and Inference for Video Segmentation Tasks} For VIS, following~\cite{vis_dataset,Cao_SipMask_ECCV_2020}, we train our network in a fair setting by randomly sampling one reference frame around the query frame. During inference, we adopt the online setting where only one previous reference frame is considered.  Since our TPR can align the feature pyramid across temporal dimension, the reference frame also contains earlier cues in a recursive manner. We adopt the same strategies for tracking heads as MaskTrack R-CNN~\cite{vis_dataset} by maintaining a memory to store the feature vectors of existing instances. For VPS, we follow the original VPS setting where we adopt UPSNet~\cite{xiong2019upsnet} with extra tracking heads~\cite{vis_dataset} as our image baseline and insert our TPR into the neck part of the model. 

\section{Experiment}

\begin{table*}[!t]
\caption{\small \textbf{Ablation studies.} We first verify the effect of each module and design choices in the first row. Then we perform several comparison and analysis of our module in the second row. Due to the limited space, more implementation details can be in supplementary.
	}\label{tab:ablations}
	\begin{minipage}[!t]{\linewidth}
	    \centering
		\begin{minipage}{.30\linewidth}
		\subfloat[Effect of two main components on various baselines.]{
		\resizebox{0.9\textwidth}{!}{%
		\centering
		\begin{tabular}{ c c c l l }
					\toprule[0.15em]
					 Backbone & + DACR  & + CPR  & mAP(\%)  \\  
					\toprule[0.15em]
    			    ResNet50 &  - & - & 33.8 \\
    			    ResNet50 &  \checkmark & - & 35.2 \\
    				ResNet50  & - & \checkmark & 34.5 \\
					ResNet50 & \checkmark & \checkmark & 36.2  \\
					\hline
					ResNet101 & - & - &  36.6 \\
					ResNet101 &\checkmark &\checkmark & 39.1 \\
					\hline
				\end{tabular}}}\hspace{3mm}
	    \end{minipage}
	    \hspace{2mm}
	    \begin{minipage}{.35\linewidth}
		\subfloat[Effect of each component in DACR. ]{
		\resizebox{0.95\textwidth}{!}{%
		\centering
		\begin{tabular}{c c c c  l}
			\toprule[0.15em]
			Method & Equ.~\ref{equ:deformable_sample} & Equ.~\ref{equ:dynaic_inner_cell_output} & Equ.~\ref{equ:dynaic_out_cell_output}& mAP(\%) \\  
		    \toprule[0.15em]
			Baseline &  &  &  & 33.8  \\ 
			& \checkmark & & & 34.5  \\
			&  & \checkmark & & 34.3  \\
			& \checkmark & \checkmark & & 34.8  \\
			&\checkmark &\checkmark &\checkmark & 35.2 \\
			\hline
		\end{tabular}	
				}}\hspace{3mm}
		\end{minipage}
		\hspace{2mm}
		\begin{minipage}{.30\linewidth}
		\subfloat[Ablation on routing space design under the same routing depth config.
		GFlops are measured on the entire validation set. 
		]{
			\footnotesize
			\resizebox{1.0\textwidth}{!}{%
			\begin{tabular}{l|c c}
				\toprule[0.15em]
				Settings & mAP(\%) & $GFlops_{avg}$   \\
				\toprule[0.15em]
				baseline+DACR &  35.2 & 58.9 \\
				\hline 
				+CPR & 36.2 &  +2.3\\
				\hline
				+Full pyramid routing & 35.0 & +6.5\\ 
				\hline
				+Full pyramid align & 33.4  & +5.9 \\
				\hline
				Top-Down routing(in CPR) & 34.6 & +4.8 \\
				\hline
		\end{tabular}}}
		\end{minipage}
	\end{minipage}
    \vspace{3pt}
	\begin{minipage}[!t]{\linewidth}
	   \centering
		\begin{minipage}{.25\linewidth}
		\subfloat[Comparison other feature warp operations.(w: with)]{
			\footnotesize
			\resizebox{0.85\textwidth}{!}{%
			\begin{tabular}{l|c }
				\toprule[0.15em]
				Sampling Method & mAP(\%)  \\
				\toprule[0.15em]
				baseline + CPR & 34.5 \\
				 w DACR  & 36.2 \\
				 \hline
				 w DCNv2 fusion~\cite{deformablev2} & 35.2 \\
			     w STSN~\cite{bertasius2018object} & 35.1 \\
			     w CBAM~\cite{woo2018cbam} & 34.8 \\
			     w optical flow warp~\cite{DFF,FlowNet} & 35.0  \\
			     w feature flow warp~\cite{yolactedge} & 34.4 \\
				\hline
		\end{tabular}}}
		\end{minipage}
        \begin{minipage}{.27\linewidth}
		\subfloat[Generalization on other instance segmentation methods.]{
			\footnotesize
			\resizebox{0.75\textwidth}{!}{%
			\begin{tabular}{l|c  }
				\toprule[0.15em]
				Settings  & mAP(\%)   \\
				\toprule[0.15em]
				SipMask~\cite{Cao_SipMask_ECCV_2020}  & 33.8  \\
				\hline
                +Our TPR & 36.2  \\
				\hline
				YOLACT~\cite{yolact-iccv2019} & 32.3   \\
				\hline
		        +Our TPR  & 34.8  \\
		       \hline
		        MaskTrack-RCNN~\cite{vis_dataset} & 30.2  \\
		        +Our TPR & 31.6  \\
				\hline
		\end{tabular}}}
	    \end{minipage}
	    \begin{minipage}{.47\linewidth}
		\subfloat[Computation Analysis on various baselines. Both GFlops and parameters are reported. GFlops are measured on the entire validation set. 
		]{
			\footnotesize
			\resizebox{1.0\textwidth}{!}{%
			\begin{tabular}{l|c c c c c c}
				\toprule[0.20em]
				Method  & Backbone & mAP(\%) & $GFlops_{avg}$ & $GFlops_{max}$ & $GFlops_{min}$ & Params \\
				\toprule[0.20em]
				Baseline & ResNet50 & 33.8 & 56.8  & 56.8 & 56.8 & 41.2M \\
				+TPR & ResNet50 & 36.2 & +4.4 & +4.9 & +4.2 & +10.2M \\
				Baseline & ResNet101 & 36.6 & 75.7 & 75.7 & 75.7 & 60.1M \\
				+TPR & ResNet101 & 39.1 & +4.9 & +6.1 & +4.5 & +14.2M \\
				\hline
		\end{tabular}}}
		\end{minipage}
	\end{minipage}
\end{table*}

\subsection{Dataset and Evaluation Metric}

\noindent
\textbf{Video Instance Segmentation}
We mainly evaluate our TPR on YouTube-VIS-2019 dataset~\cite{vis_dataset}, which has 2238 training, 302 validation, and 343 test videos. Each video is annotated with per-pixel segmentation, category, and instance ids. The dataset contains 40 object categories. Since only the validation set is available for evaluation, all results reported in this paper are evaluated on the validation set by uploading results to the online server. Following previous work, we adopt mAP  proposed in~\cite{vis_dataset} which is average precision (AP) based on a spatial-temporal Intersection-over-Union (IoU) as metrics. Both metrics adopt a COCO format where the AP is computed by averaging over multiple IoU thresholds (from 0.5 to 0.95). This metric considers all three tasks, including detection, segmentation, and tracking in a unified way. Since the dynamic network has different GFlops for different inputs, we report the average inference time (Frames Per Second, FPS) of all validation videos using one V-100 GPU card following~\cite{VIS_TR}. Following~\cite{song2020_dynamic_head}, we also report different GFlops calculated on the validation set. We also perform experiments on YouTube-VIS-2021 datasets which is an improved and augmented version of YouTube-VIS-2019 dataset. It has 2985 training videos, 421 validation videos and 453 test videos. Following~\cite{yang2021crossover}, we report our results on validation set.

\noindent
\textbf{Video Panoptic Segmentation}
We adopt the Cityscape-VPS dataset which has 400 training videos and 100 validation videos. Each video consists of 30 consecutive frames, with every 5 frames paired with the ground truth annotations. For each video, all 30 frames are predicted, and only the 6 frames with ground truth are evaluated. We adopt the Video Panoptic Quality (VPQ) as the default metric.

\begin{figure*}[!h]
	\centering
	\includegraphics[width=0.80\linewidth]{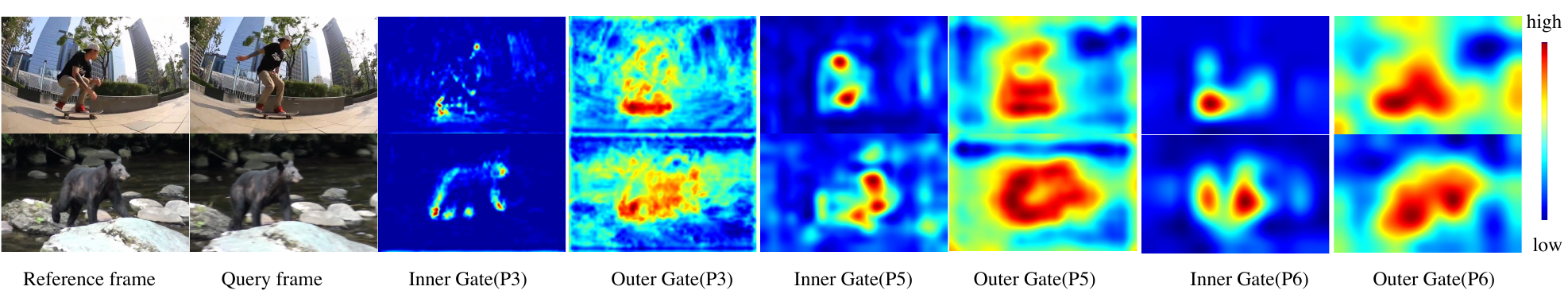}
	\caption{\small \textbf{Visualization of gates in Dynamic Aligned Cell Routing.} We choose three features (P3, P5, P6). The inner gates are all from the reference frame to control how much information is needed from the previous frame. The outer gates highlight the important regions in the current frame. Best view it on screen and zoom in. 
	}
	\label{fig:gate_res}
\end{figure*}

\subsection{Experiments on YouTube-VIS for VIS}
\label{sec:youtubevis}
\noindent
\textbf{Overview and Baseline} 
We will first perform ablation studies using BlendMask~\cite{chen2020blendmask} with ResNet50~\cite{resnet} backbone (with tracking head in Sec.~\ref{sec:nework}) as the baseline to prove the effectiveness of our TPR. Then we will give a detailed analysis and further comparison of our TPR. In addition, we also report performance gain on several other instance segmentation methods~\cite{Cao_SipMask_ECCV_2020,yolact-iccv2019,vis_dataset}. Finally, we compare our methods with previous works at last.

\noindent
\textbf{Experiment Settings}
 We use the PyTorch library~\cite{pytorch_paper} and Detectron2~\cite{detectron2} to implement all the models. Following~\cite{vis_dataset,mask_pro_vis}, all the image instance segmentation models are pre-trained on COCO datasets~\cite{COCO_dataset}. We resize the original frame size to 640$\times$360 for both training and testing. We train all the models for 12 epochs. During training, the initial learning rate is set to 0.005 and decays with a factor of 10 after epoch 8 and 11. Multiscale training is adopted to obtain a strong baseline. For each input frame, we randomly select two frames from the same video, one used as the query frame and the other used as the reference frame for training TPR and the tracking head. The reference frame is randomly sampled in a limited range of 3 for the query frame during training. All the ablation results are reported by averaging three runs for Youtube-VIS-2019.

\noindent
\textbf{Ablation study for effectiveness of TPR} We first verify the effectiveness of each component in TPR in Tab.~\ref{tab:ablations}(a). Starting from the baseline model, we obtain 33.8\% mAP. Adding DACR results in about 1.4 \% mAP gain, while CPR leads to 0.7 \% mAP, which indicates temporal cues are more important. After combining both, there is a significant gain over the baseline by 2.1 \% mAP. After using a strong baseline with ResNet101 as backbone, our TPR also has 2.5\% mAP gain, which proves the scalability of our approach. 

\noindent
\textbf{Ablation study for DACR} In Tab.~\ref{tab:ablations}(b), we investigate the necessity of
our essential modules for DACR design in Sec.~\ref{sec:tpr}. We progressively improve the baseline with: (1) DCN-based sampling in Equ.~\ref{equ:deformable_sample}; (2) dynamic routing with inner gates in Equ.~\ref{equ:dynaic_inner_cell_output}; (3) Outer gates fusion in Equ.~\ref{equ:dynaic_out_cell_output}. Consistent performance improvements can be observed after introducing the above modifications, and we can draw two essential conclusions: (1) Both DCN-based sampled and dynamic routing can align features and have the same impact. However, combining both leads to better results. (2) Outer gates work complementary with the previous two and can further improve the performance.

\noindent
\textbf{Ablation study for Cross Pyramid Routing Design} We give more routing space design comparison with our proposed Cross Pyramid Routing in Tab.~\ref{tab:ablations}(c). We conduct two different types of routing space design introduced in Sec.~\ref{sec:tpr} and Fig.~\ref{fig:align_compare}. Top-Down routing means we carry out routing from high-level features to low-level features and reverse the routing order of CPR. It results in inferior results since misalignment in low resolution is severer~\cite{2016Clockwork} and propagating such misaligned features into different scales hurts the performance for down-stream tasks. It also leads to more GFlops. The results indicate the effectiveness and efficiency of our CPR design and routing space design because all other designs will cause more misalignment across different scales and increase the ratio of misalignment parts for each feature pyramid while having more computation cost. The results verify our discussion in Sec.~\ref{sec:tpr}.

\begin{figure}[!h]
	\centering
	\includegraphics[width=1.0\linewidth]{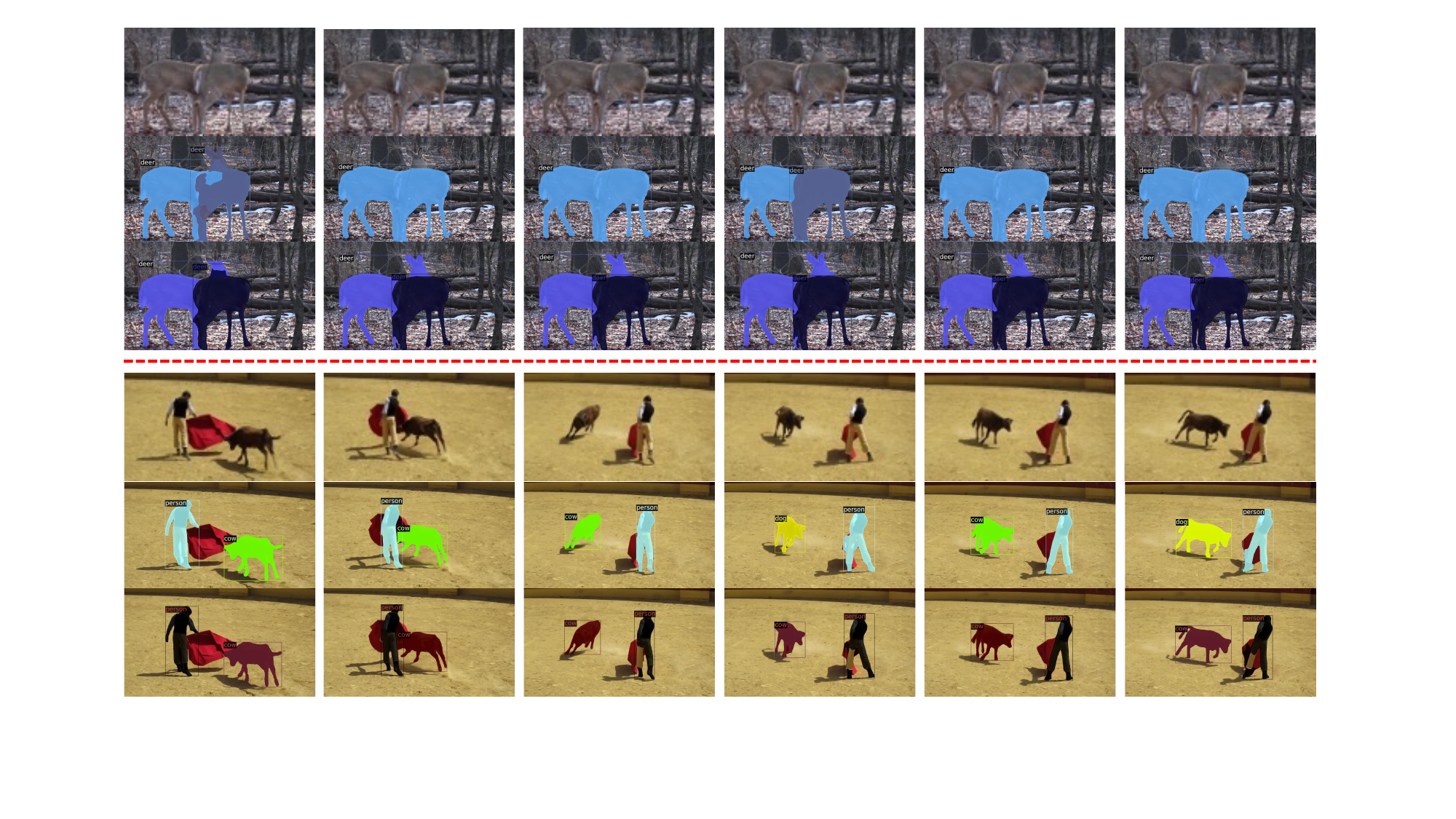}\caption{\small \textbf{Visualization results using our TPR on YouTube-VIS validation set.} Each row has five sampled frames from a video sequence. The first row for each video shows the original frames. The second row illustrates the mask predictions of Baseline Method (BlendMask with Tracking Head) and the third row those obtained with our TPR. Compared to baseline, our TPR tracks object instances more robustly even when they overlap with each other. Note that the same color represents the same object (id). Best view on the screen and zoom in.}
	\label{fig:comparison_res}
\end{figure}

\noindent
\textbf{Comparison with Warping-Based approaches} In Tab.~\ref{tab:ablations}(d), we give a further comparison with several feature warping-based approaches, including using optical flow like warping~\cite{zhu2017flowagg} and DCN-like warping~\cite{bertasius2018object}. We use the baseline method with our CPR for fair comparison. Compared with those works, our approach has the best result, mainly because our proposed double gates can well propagate more relevant information.

\noindent
\textbf{Generalization to More Instance Segmentation Approaches} We further generate our approaches on more methods, including SipMask~\cite{Cao_SipMask_ECCV_2020}, MaskTrack-RCNN~\cite{maskrcnn} and YOLACT~\cite{yolact-iccv2019}. All the methods use ResNet50 as backbone. As shown in Tab.~\ref{tab:ablations}(e), The TPR improves the results with considerable margins for different approaches. Note that all the methods are implemented in the same framework under the same setting for fair comparison.

\begin{table}[!t]\setlength{\tabcolsep}{8pt}
	\centering
	\caption{\small \textbf{Video instance segmentation} AP (\%) on the YouTube-VIS-2019~\cite{vis_dataset} validation dataset. The compared methods are listed by publication date. Note that, for the first three methods, we use the results reported by the re-implementations in~\cite{vis_dataset} for VIS. For BlendMask~\cite{chen2020blendmask} and SipMask baselines~\cite{Cao_SipMask_ECCV_2020}, we report our re-implementation results for fair comparison. \textdaggerdbl means using DCN~\cite{dai2017deformable}. The methods with underline are offline inference, while the others are online.}
	\label{tab:sota2019}
	\begin{threeparttable}
		\scalebox{0.70}{
        \begin{tabular}{ r|l|c|c|cccc}
\toprule[0.15em]
 Method&backbone& FPS& AP & $\rm AP_{50}$ & $\rm AP_{75}$ & $\rm AR_{1}$ & $\rm AR_{10}$  \\
\toprule[0.15em]
DeepSORT~\cite{wojke2017simple}&ResNet50&-&26.1&42.9&26.1&27.8&31.3  \\
FEELVOS~\cite{voigtlaender2019feelvos}&ResNet50&-&26.9&42.0&29.7&29.9&33.4  \\
OSMN~\cite{yang2018efficient}&ResNet50&-&27.5&45.1&29.1&28.6&33.1  \\
MaskTrack R-CNN~\cite{vis_dataset}&ResNet50&20.0&30.3&51.1&32.6&31.0&35.5  \\
\underline{MaskProp\cite{mask_pro_vis}}&ResNet-50\textdaggerdbl& $<2$ &40.0&-&42.9&-&-  \\
\underline{MaskProp~\cite{mask_pro_vis}}&ResNet101\textdaggerdbl & $<1$ &42.5&-&45.6&-&-  \\
\underline{STEm-Seg~\cite{Athar_Mahadevan20ECCV}}&ResNet50&-&30.6&50.7&33.5&31.6&37.1  \\
\underline{STEm-Seg~\cite{Athar_Mahadevan20ECCV}}&ResNet101&2.1&34.6&55.8&37.9&34.4&41.6  \\
CompFeat~\cite{fu2021compfeat} & ResNet50 & -  & 35.3 & 56.0 & 38.6 & 33.1 & 40.3  \\
\underline{VisTR~\cite{VIS_TR}}&ResNet50& 30.0& 36.2 & 59.8 & 36.9 & 37.2 & 42.4 \\
\underline{VisTR~\cite{VIS_TR}} &ResNet101& 27.7& 40.1 & 64.0 & 45.1 & 38.3 & 44.9  \\

\hline
{BlendMask-baseline}&ResNet50 & 19.3 & 33.4 & 52.8 & 36.1 & 32.7 & 38.0  \\
\textbf{TPR-BlendMask}&ResNet50 & 16.1 & 36.2 & 57.2 & 39.1 & 36.2 & 42.3  \\
{SipMask-baseline}&ResNet50 & 21.2 & 33.8 & 53.1 & 37.4& 35.2 & 41.0  \\
\textbf{TPR-SipMask}&ResNet50 & 15.3 & 36.0 & 57.0  & 36.8 & 36.2 & 43.6  \\
\hline
\textbf{TPR-BlendMask}&ResNet101 & 10.1 &39.1& 59.1 & 43.4 & 38.8 & 45.6  \\
\textbf{TPR-BlendMask}&ResNet101\textdaggerdbl & 8.2 & 42.2 & 63.3 & 45.7 & 40.8 & 49.0  \\
\hline
{BlendMask-baseline} & Swin-tiny & 12.3 & 38.2 & 58.1 & 42.3 & 38.3 & 44.2  \\
{BlendMask-baseline} & Swin-base & 4.7 & 44.4 & 65.2 & 48.2 & 42.1 & 50.9  \\
\textbf{TPR-BlendMask}&Swin-tiny & 10.2 &  40.0 & 62.0 & 43.4 & 38.9 & 46.7  \\
\textbf{TPR-BlendMask}&Swin-base & 3.5 & \textbf{45.9} & \textbf{67.3} &  \textbf{51.2} & \textbf{43.6} & \textbf{52.3}  \\
\hline
\end{tabular}
}

	\end{threeparttable}
\end{table}

\noindent
\textbf{Computation Analysis} Since our method is dynamic and it has different GFlops for different inputs. We report several different settings in Tab.~\ref{tab:ablations}(f) including the maximum GFlops, average GFlops and minimum GFlops.  The results prove that our TPR can not only have fewer computational overhead but also improve performance by a large margin. For instance, our method obtains about 2.1\%-2.5\% relative mAP gains over the static baseline with a lower average computational complexity (relatively only about 6.4\% - 7.7\%). That indicates the potential of our approach for the application purpose.

\noindent
\textbf{Visualization of Gate Maps} In Fig.~\ref{fig:gate_res}, we visualize several examples of our double gates. We observe that the inner gates mainly provide detailed and fine-grained instance details from the reference frame (such as heads or foot on beers) while the outer-gates focus on the current foreground objects (roughly location of instances in current frame). Both gates prohibit background noises and make the inference more efficient, which is consistent with our motivation in Sec.~\ref{equ:cell_inside}.

\noindent
\textbf{Qualitative Results} In Fig.~\ref{fig:comparison_res}, we give two visual examples of the baseline method and our TPR with each group containing images sampled from the same video. The same color represents the same instance identity. From these results, we observe that our TPR can segment instances well in two challenging situations: (1) instance overlapping~(The first group , two deers are very close to each other), (2) instance deformations and large variations~(The second groups, the appearance of the cow has changed greatly over time, thus the cow is classified into the dog wrongly.). Based on these results, we observe that our TPR reliably propagates multi-scale features that are specific to each instance with object deformations and large variations (The second example, the cow is classified into dog) and overlapping (The first example, two deers are very close with each other), in object appearance.

\noindent
\textbf{Comparison with the Previous Works on YouTube-VIS-2019} Finally, we give a detailed comparison on our methods with previous work in Tab.~\ref{tab:sota2019}. The comparison contains several aspects, including accuracy, speed, inference type and backbone network. Though MaskProp~\cite{mask_pro_vis} obtains higher mAP, that method is much slower. Moreover, MaskProp combines multiple networks such as video object detection network~\cite{bertasius2018object}, Hybrid Task Cascade Network~\cite{htc} for cascade feature learning and the complex High-Resolution Mask Refinement post-processing. Our method shares a much simpler pipeline and performs the online inference. As shown in Tab.~\ref{tab:sota2019}, our TPR improves the various models with little FPS drop. In particular, with ResNet101 and DCN~\cite{dai2017deformable} as backbone, our method can achieve \textbf{42.2\% mAP} while running at 8.2 FPS. We also verify the generality on larger Swin Transformer~\cite{liu2021swin} as backbone in the bottom of Tab.~\ref{tab:sota2019}. After using the Swin-base backbone, our method can achieve 45.9 AP while running at 4.8 FPS which is better and faster than the MaskProp.

\noindent
\textbf{Comparison with the Previous Works on YouTube-VIS-2021}
We report the recently proposed MaskTrack R-CNN and SipMask-VIS on this dataset using official implementation for comparison. We also compare the recently proposed CrossVIS~\cite{yang2021crossover} for fair comparison. From the Tab.~\ref{tab:sota_vis_2021}, our method achieves better results than previous works.

\begin{table}[!t]\setlength{\tabcolsep}{8pt}
	\centering
	\caption{\small \textbf{Video instance segmentation} AP (\%) on the new YouTube-VIS-2021~\cite{vis_dataset} validation dataset.}
	\label{tab:sota_vis_2021}
	\begin{threeparttable}
		\scalebox{0.80}{
        \begin{tabular}{ r|c|c|cccc}
\toprule[0.15em]
 Method&backbone& AP & $\rm AP_{50}$ & $\rm AP_{75}$ & $\rm AR_{1}$ & $\rm AR_{10}$  \\
\toprule[0.15em]

MaskTrack R-CNN~\cite{vis_dataset}&ResNet50&30.3&51.1&32.6&31.0&35.5  \\
SipMask-VIS~\cite{Cao_SipMask_ECCV_2020}&ResNet50&31.7&52.5&34.0&30.8&37.8  \\
BlendMask-VIS~\cite{chen2020blendmask}&ResNet50&32.5&53.3&35.0&31.8&38.8  \\
Cross-VIS~\cite{yang2021crossover} &ResNet50& 34.2 & 54.4 & 37.9 & 30.4 & 38.2 \\
\hline
\textbf{TPR-BlendMask}&ResNet50 & 35.3 & 54.8 & 37.9 & 34.2 & 43.6  \\
\textbf{TPR-BlendMask}&ResNet101 &38.1& 59.1 & 43.4 & 38.8 & 45.6  \\
\hline
\end{tabular}
}

	\end{threeparttable}
\end{table}

\begin{table}[!t]\setlength{\tabcolsep}{8pt}
	\centering
	\caption{\small Comparisons on Cityscapes-VPS dataset. We report our reproduced baseline in ~\cite{kim2020vps}.}
	\label{tab:comparison_on_vps}
	\begin{threeparttable}
		\scalebox{1.0}{
			\begin{tabular}{ r|c|cc}
				\toprule[0.15em]
				Method& VPQ & $VPQ_{thing}$ & $VPQ_{stuff}$ \\
				\toprule[0.15em]
				UPSNet+Track~\cite{vis_dataset}& 55.8 & 43.9 & 65.3 \\
				\textbf{+TPR} & 56.7 &  44.5 &  66.1  \\
				\hline
				UPSNet+Track+Fuse~\cite{kim2020vps}& 56.8 & 44.2 & 66.4 \\
				\textbf{+TPR} & 57.5 & 44.7 & 66.8  \\
				\hline	
			\end{tabular}
		}
	\end{threeparttable}
\end{table}

\subsection{Experiments on Cityscapes-VPS for VPS}
In Tab.~\ref{tab:comparison_on_vps}, we compare our methods with strong UPSNet with track head as baseline on Cityscapes-VPS dataset. Our method can achieve significant gain on both thing class and stuff class. This proves the potential of TPR for other video scene understanding tasks.

\section{Conclusion}
In this paper, we propose a conceptually new framework named Temporal Pyramid Routing (TPR) for dynamic instance learning in video. The pyramid features are well aligned and routed along the temporal dimension via TPR. In particular, we design specific Dynamic Aligned Cell Routing (DACR) with a double gate design to avoid background noise while capturing useful information during the propagation. Moreover, the Cross Pyramid Routing (CPR) is proposed to propagate aligned features into all scales. Extensive experiments demonstrate the effectiveness and efficiency of our TPR on VIS and VPS task. We believe this work can provide several insights into future works for modeling dynamics in video in the aspect of video dynamic network design.

\ifCLASSOPTIONcaptionsoff
  \newpage
\fi



{
\bibliographystyle{IEEEtran}
\bibliography{IEEEabrv,egbib}
}

\end{document}


%
\title{Improving Video Instance Segmentation via Temporal Pyramid Routing-Appendix}
%
%
%
%

\author{Xiangtai Li,
        Hao He,
        Yibo Yang,
        Henghui Ding,
        Kuiyuan Yang,
        Guangliang Cheng,\\
        Yunhai Tong \textrm{\Letter},
        Dacheng Tao,~\textit{Fellow}, IEEE
        
\IEEEcompsocitemizethanks{\IEEEcompsocthanksitem X. Li and Y. Tong are with the School of Electronics Engineering and Computer Science, Peking Univeristy, Beijing, China. This work is supported by the National Key Research and Development Program of China (No.2020YFB2103402). 
\IEEEcompsocthanksitem H.~He is with the Department of National Laboratory of Pattern Recognition, Institute of Automation, Beijing , China.
\IEEEcompsocthanksitem Y.~Yang and D.~Tao are with JD Explore Academy, Beijing, China.
\IEEEcompsocthanksitem G.~Cheng is with SenseTime Research, Beijing, China.
\IEEEcompsocthanksitem H.~Ding is with ETH Zurich, Switzerland.
\IEEEcompsocthanksitem K.~Yang is with Xiaomi, Beijing, China.
}
}

%
%

\markboth{IEEE TRANSACTIONS ON PATTERN ANALYSIS AND MACHINE INTELLIGENCE,~Vol.~X, No.~X, X}
{Shell \MakeLowercase{\textit{et al.}}: Bare Advanced Demo of IEEEtran.cls for IEEE Computer Society Journals}
%



\IEEEtitleabstractindextext{


}

\maketitle

\IEEEdisplaynontitleabstractindextext

%
\IEEEpeerreviewmaketitle

\noindent
\textbf{Overview.} This supplementary mainly contains three parts. The first part describes the implementation details for the experiment section in the main paper. The second parts present more ablation studies on the key design of TPR including DACR and loss design. The third parts give more visualization results of our proposed methods. 

\section{More Experiments Details}
\label{sec:more_exp}

\noindent
\textbf{Implementation Details of Instance Segmentation Methods.} 
For fair comparison, we adopt the same detector FCOS~\cite{tian2021fcos} for all the single stage instance segmentation methods including SipMask~\cite{Cao_SipMask_ECCV_2020}, YOLACT~\cite{yolact-iccv2019} and BlendMask~\cite{chen2020blendmask}. For ResNet50 backbone, we pre-train all the models on coco dataset~\cite{COCO_dataset} for 12 epochs. For ResNet101 backbone, we pre-train all the models for 36 epochs and the pre-trained models are used as the stronger baselines.

\noindent
\textbf{Choice of Depth configuration in CPR.}
By default, we adopt the depth configuration as $[4,2,1,1]$. In Tab~\ref{tab:influnce_depth_config}, we try other settings where we found the first stage routing plays a more important role for the final performance. That indicates the fine-grained features are more aligned and lead to better representation. Then we increase the depth in remaining stages where we found no gain or even worse results. We argue that introducing more low-resolution features leads to bad results on TPR since they are not well aligned. The results are consistent with previous work~\cite{2016Clockwork}. 

\noindent
\textbf{Effect of gating in DACR.} In Tab.~\ref{tab:gating_dacr}, we remove the gating during the inference, which means all pixels are involved. We find about 1.1\% mAP drop. This verifies that our dynamic design makes sparse alignment and avoids the background noise.

\noindent
\textbf{Effect of $\lambda1$ for Budget Loss.} In Tab.~\ref{tab:effect_of_budge_loss}, we perform ablation on $\lambda1$ of the budget loss. Increasing $\lambda1$ results in significant performance drops since no extra temporal information is propagated into the next frame.

\noindent
\textbf{Effect of inner gates and outer gates.} In Tab.~\ref{tab:DCN_only},
we preform more ablation studies on the effect of inner gates and outer gates on various works using Blendmask baseline. We find

\begin{table}[!t]\setlength{\tabcolsep}{6pt}
	\centering
	\begin{threeparttable}
		\scalebox{0.95}{
			\begin{tabular}{l c }
	\toprule[0.1em]
	Settings  & mAP \\
	\toprule[0.1em]
    baseline+DACR & 35.2 \\
    \hline
     $[4,2,1,1]$ (default) & 35.9  \\
    \hline
    $[3,1,1,1]$ & 35.4 \\
    $[4,1,1,1]$ & 35.6 \\
    \hline
   $[1,4,1,1]$ & 35.0 \\
   $[1,1,4,1]$ & 35.1 \\
   $[1,1,1,4]$ & 34.9 \\
   \hline
   $[4,2,2,1]$  & 35.3  \\
   $[4,2,2,2]$  & 35.0  \\
	\bottomrule[0.1em]
	\end{tabular}}
		\caption{Influence of the Depth Configuration in CPR.}
		\label{tab:influnce_depth_config}
	\end{threeparttable}
\end{table}

\begin{table}[!t]\setlength{\tabcolsep}{6pt}
	\centering
	\begin{threeparttable}
		\scalebox{0.95}{
			\begin{tabular}{l c }
	\toprule[0.1em]
	Settings  & mAP \\
	\toprule[0.1em]
    our TPR & 36.2 \\
    \hline
    our TPR w/o gates in DACR & 35.3  \\

	\bottomrule[0.1em]
	\end{tabular}}
		\caption{Effect of gating in DACR.}
		\label{tab:gating_dacr}
	\end{threeparttable}
\end{table}

\begin{table}[!t]\setlength{\tabcolsep}{6pt}
	\centering
	\begin{threeparttable}
		\scalebox{0.95}{
			\begin{tabular}{l c }
	\toprule[0.1em]
	Settings  & mAP \\
	\toprule[0.1em]
     r50 + TPR (baseline) & 35.2  \\
     removing gates & 34.5 \\
     \hline
          r101 + TPR (baseline) & 39.1  \\
     removing gates & 37.0 \\
     \hline
          Swin-Tiny + TPR (baseline) & 40.0  \\
     removing gates & 38.4 \\
	\bottomrule[0.1em]
	\end{tabular}}
		\caption{Effect of removing inner gates and outer gates.}
		\label{tab:DCN_only}
	\end{threeparttable}
\end{table}

\begin{table}[!t]\setlength{\tabcolsep}{6pt}
	\centering
	\begin{threeparttable}
		\scalebox{0.95}{
			\begin{tabular}{l c }
	\toprule[0.1em]
	Settings  & mAP \\
	\toprule[0.1em]
    baseline+DACR & 35.2 \\
    \hline
     $\lambda_1 =1.0$ &  36.0  \\
     $\lambda_1 =1.5$ (default) &  36.2  \\
     $\lambda_1 =2.5$  & 35.2 \\
     $\lambda_1 =5.5$  & 34.1 \\
	\bottomrule[0.1em]
	\end{tabular}}
		\caption{Effect of $\lambda_1$ for budget loss.}
		\label{tab:effect_of_budge_loss}
	\end{threeparttable}
\end{table}

\section{More Visualization Results}
\label{sec:vis_res}

\begin{figure*}[!t]
	\centering
	\includegraphics[width=1.0\linewidth]{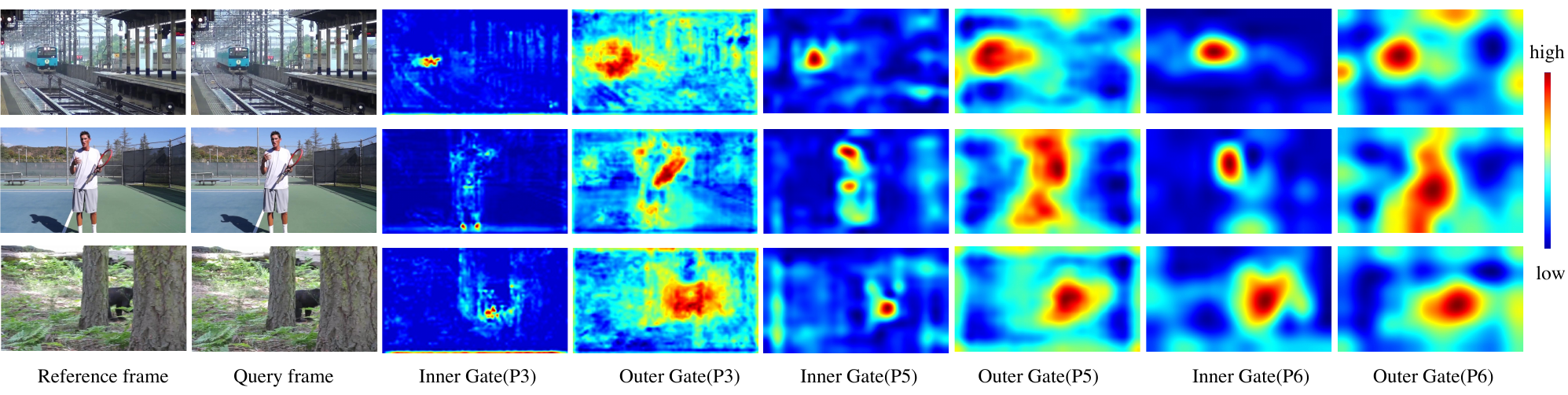}
	\caption{\textbf{More Visualization of gates in Dynamic Aligned Cell Routing.} We choose three features (P3, P5, P6). The inner gates are all from the reference frame to control how much information is needed from the previous frame. The outer gates highlight the important regions in the current frame. Best view it on screen and Zoom in. 
	}
	\label{fig:gate_res_supp}
\end{figure*}

\noindent
\textbf{Visualization of Learned Gates.}
In Fig.~\ref{fig:gate_res_supp}, we give more visualization of dynamic learned gate maps in DACR. We observe the same results as the main paper, which verify our motivation: inner gates mainly provide detailed and fine-grained instance details from the reference frame, while the outer-gates focus on the current foreground objects. The gates make the inference process more efficient.

\begin{figure*}[!h]
	\centering
	\includegraphics[width=0.80\linewidth]{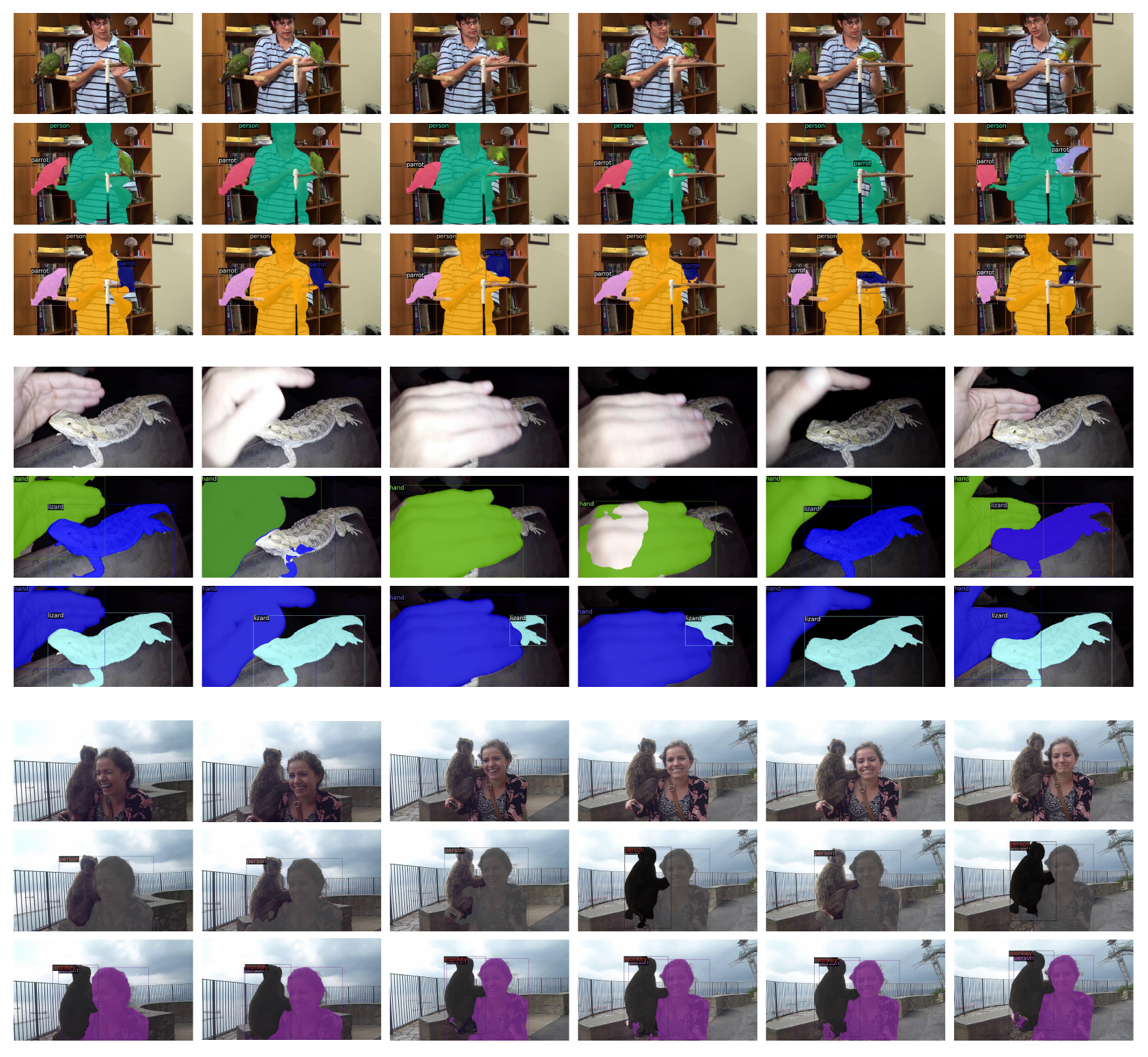}
	\caption{\textbf{More Visualization results using our TPR on YouTube-VIS validation set.} Each row has six sampled frames from a video sequence. The first row for each video shows the original frames. The second row illustrates the mask predictions of Baseline Method(BlendMask with Tracking Head) and the third row those obtained with our TPR. Compared to baseline, our TPR tracks object instances more robustly even when they overlap with each other. Note that the same color represents the same object(id). Best view on the screen and zoom in.}
	\label{fig:comparison_res_supp}
\end{figure*}

\noindent
\textbf{More Comparison Results.}
In Fig.~\ref{fig:comparison_res_supp}, we give more visual examples of the baseline method and our TPR, with each group containing images sampled from the same video.

\ifCLASSOPTIONcaptionsoff
  \newpage
\fi



{
\bibliographystyle{IEEEtran}
\bibliography{IEEEabrv,egbib}
}